\documentclass[english]{article}
\usepackage[T1]{fontenc}
\usepackage[latin9]{inputenc}
\usepackage{babel}
\usepackage{float}
\usepackage{units}
\usepackage{amsmath}
\usepackage{amssymb}
\usepackage{graphicx}
\usepackage[unicode=true]
 {hyperref}

\makeatletter

\providecommand{\tabularnewline}{\\}

\usepackage[final,nonatbib]{neurips_2018}
\renewcommand{\textendash}{--}

\makeatother

\begin{document}

\title{Deep Generative Markov State Models}

\author{Hao Wu$^{1,2,*}$, Andreas Mardt$^{1,*}$, Luca Pasquali$^{1,}$\thanks{H.~Wu, A.~Mardt and L.~Pasquali equally contributed to this work.}~ and Frank No{\'e}$^{1,}$\thanks{Author to whom correspondence should be addressed. Electronic mail: frank.noe@fu-berlin.de.}\\$^1$Dept. of Mathematics and Computer Science, Freie Universität Berlin, Arnimallee 6, 14195 Berlin, Germany\\$^2$School of Mathematical Sciences, Tongji University, Shanghai, 200092, P.R.~China}
\maketitle
\begin{abstract}
We propose a deep generative Markov State Model (DeepGenMSM) learning
framework for inference of metastable dynamical systems and prediction
of trajectories. After unsupervised training on time series data,
the model contains (i) a probabilistic encoder that maps from high-dimensional
configuration space to a small-sized vector indicating the membership
to metastable (long-lived) states, (ii) a Markov chain that governs
the transitions between metastable states and facilitates analysis
of the long-time dynamics, and (iii) a generative part that samples
the conditional distribution of configurations in the next time step.
The model can be operated in a recursive fashion to generate trajectories
to predict the system evolution from a defined starting state and
propose new configurations. The DeepGenMSM is demonstrated to provide
accurate estimates of the long-time kinetics and generate valid distributions
for molecular dynamics (MD) benchmark systems. Remarkably, we show
that DeepGenMSMs are able to make long time-steps in molecular configuration
space and generate physically realistic structures in regions that
were not seen in training data.
\end{abstract}

\section{Introduction}

Complex dynamical systems that exhibit events on vastly different
timescales are ubiquitous in science and engineering. For example,
molecular dynamics (MD) of biomolecules involve fast vibrations on
the timescales of $10^{-15}$ seconds, while their biological function
is often related to the rare switching events between long-lived states
on timescales of $10^{-3}$ seconds or longer. In weather and climate
systems, local fluctuations in temperature and pressure fields occur
within minutes or hours, while global changes are often subject to
periodic motion and drift over years or decades. Primary goals in
the analysis of complex dynamical systems include: 
\begin{enumerate}
\item \emph{Deriving an interpretable model} of the essential long-time
dynamical properties of these systems, such as the stationary behavior
or lifetimes/cycle times of slow processes.
\item \emph{Simulating the dynamical system}, e.g., to predict the system's
future evolution or to sample previously unobserved system configurations.
\end{enumerate}
A state-of-the-art approach for the first goal is to learn a Markovian
model from time-series data, which is theoretically justified by the
fact that physical systems are inherently Markovian. In practice,
the long-time behavior of dynamical systems can be accurately described
in a Markovian model when suitable features or variables are used,
and when the time resolution of the model is sufficiently coarse such
that the time-evolution can be represented with a manageable number
of dynamical modes \cite{SarichNoeSchuette_MMS09_MSMerror,KordaMezic_JNLS2017_ConvergenceEDMD}.
In stochastic dynamical systems, such as MD simulation, variants of
Markov state models (MSMs) are commonly used \cite{BowmanPandeNoe_MSMBook,SarichSchuette_MSMBook13,PrinzEtAl_JCP10_MSM1}.
In MSMs, the configuration space is discretized, e.g., using a clustering
method, and the dynamics between clusters are then described by a
matrix of transition probabilities \cite{PrinzEtAl_JCP10_MSM1}. The
analogous approach for deterministic dynamical systems such as complex
fluid flows is called Koopman analysis, where time propagation is
approximated by a linear model in a suitable function space transformation
of the flow variables \cite{Mezic_NonlinDyn05_Koopman,SchmidSesterhenn_APS08_DMD,TuEtAl_JCD14_ExactDMD,BruntonProctorKutz_PNAS16_Sindy}.
The recently proposed VAMPnets learn an optimal feature transformation
from full configuration space to a low-dimensional latent space in
which the Markovian model is built by variational optimization of
a neural network \cite{MardtEtAl_VAMPnets}. When the VAMPnet has
a probabilistic output (e.g. SoftMax layer), the Markovian model conserves
probability, but is not guaranteed to be a valid transition probability
matrix with nonnegative elements. A related work for deterministic
dynamical systems is Extended Dynamic Mode Decomposition with dictionary
learning \cite{LiKevekidis_EDMD-DL}. All of these methods are purely
analytic, i.e. they learn a reduced model of the dynamical system
underlying the observed time series, but they miss a generative part
that could be used to sample new time series in the high-dimensional
configuration space.

Recently, several learning frameworks for dynamical systems have been
proposed that partially address the second goal by including a decoder
from the latent space back to the space of input features. Most of
these methods primarily aim at obtaining a low-dimensional latent
space that encodes the long-time behavior of the system, and the decoder
takes the role of defining or regularizing the learning problem \cite{WehmeyerNoe_TAE,HernandezPande_VariationalDynamicsEncoder,LuschKutzBrunton_DeepKoopman,OttoRowley_LinearlyRecurrentAutoencoder,RibeiroTiwary_JCP18_RAVE}.
In particular none of these models have demonstrated the ability to
generate viable structures in the high-dimensional configuration space,
such as a molecular structure with realistic atom positions in 3D.
Finally, some of these models learn a linear model of the long-timescale
dynamics \cite{LuschKutzBrunton_DeepKoopman,OttoRowley_LinearlyRecurrentAutoencoder},
but none of them provide a probabilistic dynamical model that can
be employed in a Bayesian framework. Learning the correct long-time
dynamical behavior with a generative dynamical model is difficult,
as demonstrated in \cite{HernandezPande_VariationalDynamicsEncoder}.

Here, we address these aforementioned gaps by providing a deep learning
framework that learns, based on time-series data, the following components:
\begin{enumerate}
\item Probabilistic encodings of the input configuration to a low-dimensional
latent space by neural networks, $x_{t}\rightarrow\boldsymbol{\chi}(x_{t})$.
\item A true transition probability matrix $\mathbf{K}$ describing the
system dynamics in latent space for a fixed time-lag $\tau$:
\[
\mathbb{E}\left[\boldsymbol{\chi}(x_{t+\tau})\right]=\mathbb{E}\left[\mathbf{K}^{\top}(\tau)\boldsymbol{\chi}(x_{t})\right].
\]
The probabilistic nature of the method allows us to train it with
likelihood maximization and embed it into a Bayesian framework. In
our benchmarks, the transition probability matrix approximates the
long-time behavior of the underlying dynamical system with high accuracy.
\item A generative model from latent vectors back to configurations, allowing
us to sample the transition density $\mathbb{P}(x_{t+\tau}|x_{t})$
and thus propagate the model in configuration space. We show for the
first time that this allows us to sample genuinely new and valid molecular
structures that have not been included in the training data. This
makes the method promising for performing active learning in MD \cite{BowmanEnsignPande_JCTC2010_AdaptiveSampling,PlattnerEtAl_NatChem17_BarBar},
and to predict the future evolution of the system in other contexts.
\end{enumerate}

\section{Deep Generative Markov State Models}

\begin{figure}
\begin{centering}
\includegraphics[width=0.7\columnwidth]{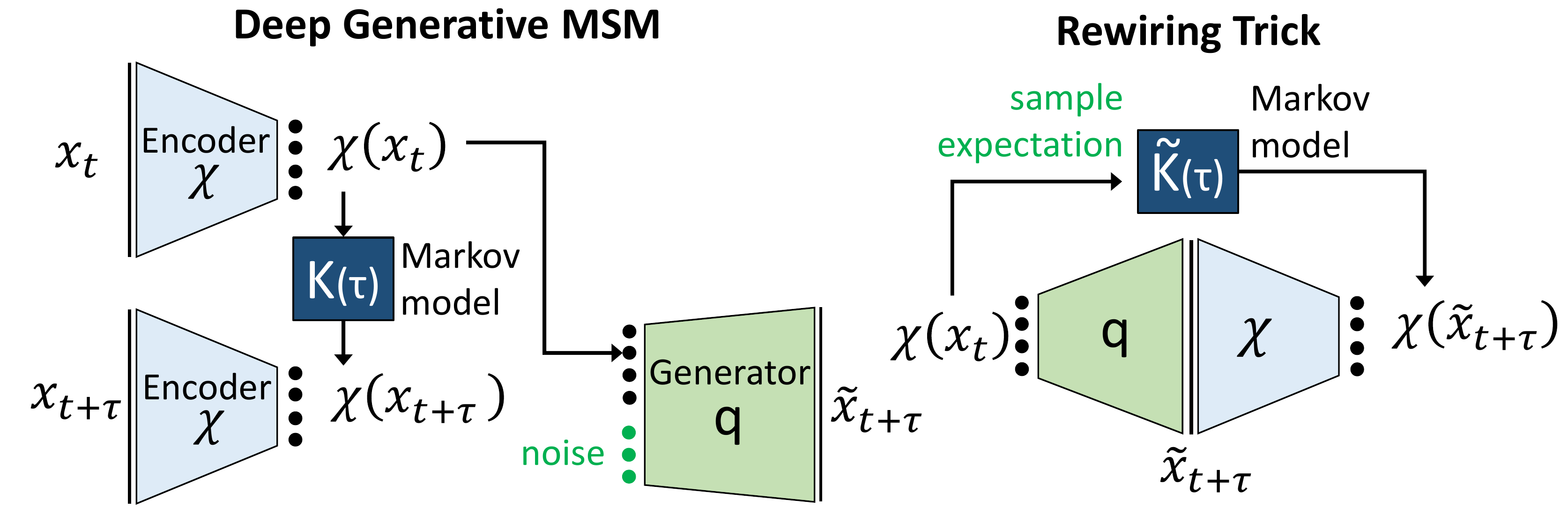}
\par\end{centering}
\caption{\label{fig:scheme}Schematic of Deep Generative Markov State Models
(DeepGenMSMs) and the rewiring trick. The function $\chi$, here represented
by neural networks, maps the time-lagged input configurations to metastable
states whose dynamics are governed by a transition probability matrix
$\mathbf{K}$. The generator samples the distribution $x_{t+\tau}\sim\mathbf{q}$
by employing a generative network that can produce novel configurations
(or by resampling $x_{t+\tau}$ in DeepResampleMSMs). The rewiring
trick consists of reconnecting the probabilistic networks $\mathbf{q}$
and $\boldsymbol{\chi}$ such that the time propagation in latent
space can be sampled: From the latent state $\chi(x_{t})$, we generate
a time-lagged configuration $x_{t+\tau}$ using $\mathbf{q}$, and
then transform it back to the latent space, $\chi(x_{t+\tau})$. Each
application of the rewired network samples the latent space transitions,
thus providing the statistics to estimate the Markov model transition
matrix $\mathbf{K}(\tau)$, which is needed for analysis. This trick
allows $\mathbf{K}(\tau)$ to be estimated with desired constraints,
such as detailed balance.}
\end{figure}

Given two configurations $x,y\in\mathbb{R}^{d}$, where $\mathbb{R}^{d}$
is a potentially high-dimensional space of system configurations (e.g.
the positions of atoms in a molecular system), Markovian dynamics
are defined by the transition density $\mathbb{P}(x_{t+\tau}=y|x_{t}=x)$.
Here we represent the transition density between $m$ states in the
following form (Fig. \ref{fig:scheme}):
\begin{equation}
\mathbb{P}(x_{t+\tau}=y|x_{t}=x)=\boldsymbol{\chi}(x)^{\top}\mathbf{q}(y;\tau)=\sum_{i=1}^{m}\chi_{i}(x)q_{i}(y;\tau).\label{eq:Ansatz}
\end{equation}
Here, $\boldsymbol{\chi}(x)^{\top}=[\chi_{1}(x),...,\chi_{m}(x)]$
represent the probability of configuration $x$ to be in a metastable
(long-lived) state $i$
\begin{align*}
\chi_{i}(x) & =\mathbb{P}(x_{t}\in\text{state }i\mid x_{t}=x).
\end{align*}
Consequently, these functions are nonnegative ($\chi_{i}(x)\ge0\:\:\forall x$)
and sum up to one ($\sum_{i}\chi_{i}(x)=1\:\:\forall x$). The functions
$\boldsymbol{\chi}(x)$ can, e.g., be represented by a neural network
mapping from $\mathbb{R}^{d}$ to $\mathbb{R}^{m}$ with a SoftMax
output layer. Additionally, we have the probability densities
\[
q_{i}(y;\tau)=\mathbb{P}(x_{t+\tau}=y|x_{t}\in\text{state }i)
\]
that define the probability density of the system to ``land'' at
configuration $y$ after making one time-step. We thus briefly call
them ``landing densities''.

\subsection{Kinetics}

Before addressing how to estimate $\boldsymbol{\chi}$ and $\mathbf{q}$
from data, we describe how to perform the standard calculations and
analyses that are common in the Markov modeling field for a model
of the form (\ref{eq:Ansatz}).

In Markov modeling, one is typically interested in the kinetics of
the system, i.e. the long-time behavior of the dynamics. This is captured
by the elements of the transition matrix $\mathbf{K}=[k_{ij}]$ between
metastable states. $\mathbf{K}$ can be computed as follows: the product
of the probability density to jump from metastable $i$ to a configuration
$y$ and the probability that this configuration belongs to metastable
state $j$, integrated over the whole configuration space.
\begin{equation}
k_{ij}(\tau)=\int_{y}q_{i}(y;\tau)\chi_{j}(y)\,\mathrm{d}y.\label{eq:transition_matrix}
\end{equation}
Practically, this calculation is implemented via the ``rewiring trick''
shown in Fig. \ref{fig:scheme}, where the configuration space integral
is approximated by drawing samples from the generator. The estimated
probabilistic functions $\mathbf{q}$ and $\boldsymbol{\chi}$ define,
by construction, a valid transition probability matrix $\mathbf{K}$,
i.e. $k_{ij}\ge0$ and $\sum_{j}k_{ij}=1$. As a result, the proposed
models  have a structural advantage over other high-accuracy Markov
state modeling approaches that define metastable states in a fuzzy
or probabilistic manner but do not guarantee a valid transition matrix
\cite{KubeWeber_JCP07_CoarseGraining,MardtEtAl_VAMPnets} (See Supplementary
Material for more details.).

The stationary (equilibrium) probabilities of the metastable states
are given by the vector $\boldsymbol{\pi}=[\pi_{i}]$ that solves
the eigenvalue problem with eigenvalue $\lambda_{1}=1$:
\begin{equation}
\boldsymbol{\pi}=\mathbf{K}^{\top}\boldsymbol{\pi},\label{eq:stationary_vector}
\end{equation}
and the stationary (equilibrium) distribution in configuration space
is given by:
\begin{equation}
\mu(y)=\sum_{i}\pi_{i}q_{i}(y;\tau)=\boldsymbol{\pi}^{\top}\mathbf{q}(y;\tau).\label{eq:stationary_density}
\end{equation}
Finally, for a fixed definition of states \emph{via} $\boldsymbol{\chi}$,
the self-consistency of Markov models may be tested using the Chapman-Kolmogorov
equation
\begin{equation}
\mathbf{K}^{n}(\tau)\approx\mathbf{K}(n\tau)\label{eq:CKtest}
\end{equation}
which involves estimating the functions $\mathbf{q}(y;n\tau)$ at
different lag times $n\tau$ and comparing the resulting transition
matrices with the $n$th power of the transition matrix obtained at
lag time $\tau$. A consequence of Eq. (\ref{eq:CKtest}) is that
the relaxation times
\begin{equation}
t_{i}(\tau)=-\frac{\tau}{\log|\lambda_{i}(\tau)|}\label{eq:relaxation_time}
\end{equation}
are independent of the lag time $\tau$ at which $\mathbf{K}$ is
estimated \cite{SwopePiteraSuits_JPCB108_6571}. Here, $\lambda_{i}$
with $i=2,...,m$ are the nontrivial eigenvalues of $\mathbf{K}$.

\subsection{Maximum Likelihood (ML) learning of DeepResampleMSM}

Given trajectories $\{x_{t}\}_{t=1,...,T}$, how do we estimate the
membership probabilities $\boldsymbol{\chi}(x)$, and how do we learn
and sample the landing densities $\mathbf{q}(y)$? We start with a
model, where $\mathbf{q}(y)$ are directly derived from the observed
(empirical) observations, i.e. they are point densities on the input
configurations $\{x_{t}\}$, given by:
\begin{equation}
q_{i}(y)=\frac{1}{\bar{\gamma}_{i}}\gamma_{i}(y)\rho(y).\label{eq:vamp-model}
\end{equation}
Here, $\rho(y)$ is the empirical distribution, which in the case
of finite sample size is simply $\rho(y)=\frac{1}{T-\tau}\sum_{t=1}^{T-\tau}\delta(y-x_{t+\tau})$,
and $\gamma_{i}(y)$ is a trainable weighting function. The normalization
factor  $\bar{\gamma}_{i}=\frac{1}{T-\tau}\sum_{t=1}^{T-\tau}\gamma_{i}(x_{t+\tau})=\mathbb{E}_{y\sim\rho_{1}}[\gamma_{i}(y)]$
ensures $\int_{y}q_{i}(y)\,\mathrm{d}y=1$.

Now we can optimize $\chi_{i}$ and $\gamma_{i}$ by maximizing the
likelihood (ML) of generating the pairs $(x_{t},x_{t+\tau})$ observed
in the data. The log-likelihood is given by:
\begin{equation}
LL=\sum_{t=1}^{T-\tau}\ln\left(\sum_{i=1}^{m}\chi_{i}(x_{t})\bar{\gamma}_{i}^{-1}\gamma_{i}(x_{t+\tau})\right),\label{eq:LL}
\end{equation}
and is maximized to train a deep MSM with the structure shown in Fig.
\ref{fig:scheme}.

Alternatively, we can optimize $\chi_{i}$ and $\gamma_{i}$ using
the Variational Approach for Markov Processes (VAMP) \cite{WuNoe_VAMP}.
However, we found the ML approach to perform significantly better
in our tests, and we thus include the VAMP training approach only
in the Supplementary Material without elaborating on it further. 

Given the networks $\boldsymbol{\chi}$ and $\boldsymbol{\gamma}$,
we compute $\mathbf{q}$ from Eq. (\ref{eq:vamp-model}). Employing
the rewiring trick shown in Fig. \ref{fig:scheme} results in computing
the transition matrix by a simple average over all configurations:
\begin{equation}
\mathbf{K}=\frac{1}{N}\sum_{t=\tau}^{T-\tau}\mathbf{q}(x_{t+\tau})\boldsymbol{\chi}(x_{t+\tau})^{\top}.
\end{equation}
The deep MSMs described in this section are neural network generalizations
of traditional MSMs \textendash{} they learn a mapping from configurations to
metastable states, where they aim obtaining a good approximation of
the kinetics of the underlying dynamical system, by means of the transition
matrix $\mathbf{K}$. However, since the landing distribution $\mathbf{q}$
in these methods is derived from the empirical distribution (\ref{eq:vamp-model}),
any generated trajectory will only resample configurations from the
input data. To highlight this property, we will refer to the deep
MSMs with the present methodology as DeepResampleMSM. 

\subsection{Energy Distance learning of DeepGenMSM}

In contrast to DeepResampleMSM, we now want to learn deep generative
MSM (DeepGenMSM), which can be used to generate trajectories that
do not only resample from input data, but can produce genuinely new
configurations. To this end, we train a generative model to mimic
the empirical distribution $q_{i}(y)$:
\begin{equation}
y=G(e_{i},\epsilon),\label{eq:G}
\end{equation}
where the vector $e_{i}\in\mathbb{R}^{m}$ is a one-hot encoding of
the metastable state, and $\epsilon$ is a i.i.d. random vector where
each component samples from a Gaussian normal distribution. 

Here we train the generator $G$ by minimizing the conditional Energy
Distance (ED), whose choice is motivated in the Supplementary Material.
The standard ED, introduced in \cite{SzekelyRizzo_Interstat04_EnergyDistance},
is a metric between the distributions of random vectors, defined as
\begin{equation}
D_{E}\left(\mathbb{P}(\mathbf{x}),\mathbb{P}(\mathbf{y})\right)=\mathbb{E}\left[2\left\Vert x-y\right\Vert -\left\Vert x-x^{\prime}\right\Vert -\left\Vert y-y^{\prime}\right\Vert \right]
\end{equation}
for two real-valued random variables $\mathbf{x}$ and $\mathbf{y}$.
$x,x^{\prime},y,y^{\prime}$ are independently distributed according
to the distributions of $\mathbf{x},\mathbf{y}$. Based on this metric,
we introduce the conditional energy distance between the transition
density of the system and that of the generative model:
\begin{eqnarray}
D & \triangleq & \mathbb{E}\left[D_{E}\left(\mathbb{P}(\mathbf{x}_{t+\tau}|x_{t}),\mathbb{P}(\hat{\mathbf{x}}_{t+\tau}|x_{t})\right)|x_{t}\right]\nonumber \\
 & = & \mathbb{E}\left[2\left\Vert \hat{x}_{t+\tau}-x_{t+\tau}\right\Vert -\left\Vert \hat{x}_{t+\tau}-\hat{x}_{t+\tau}^{\prime}\right\Vert -\left\Vert x_{t+\tau}-x_{t+\tau}^{\prime}\right\Vert \right]\label{eq:conditional-energy-distance}
\end{eqnarray}
Here $x_{t+\tau}$ and $x_{t+\tau}^{\prime}$ are distributed according
to the transition density for given $x_{t}$ and $\hat{x}_{t+\tau},\hat{x}_{t+\tau}^{\prime}$
are independent outputs of the generative model conditioned on $x_{t}$.
Implementing the expectation value with an empirical average results
in an estimate for $D$ that is unbiased, up to an additive constant.
We train $G$ to minimize $D$. See Supplementary Material for detailed
derivations and the training algorithm used.

After training, the transition matrix can be obtained by using the
rewiring trick (Fig. \ref{fig:scheme}), where the configuration space
integral is sampled by generating samples from the generator:
\begin{equation}
\left[\mathbf{K}\right]_{ij}=\mathbb{E}_{\epsilon}\left[\chi_{j}\left(G(e_{i},\epsilon)\right)\right].
\end{equation}

\section{Results}

Below we establish our framework by applying it to two well-defined
benchmark systems that exhibit metastable stochastic dynamics. We
validate the stationary distribution and kinetics by computing $\boldsymbol{\chi}(x)$,
$\mathbf{q}(y)$, the stationary distribution $\mu(y)$ and the relaxation
times $t_{i}(\tau)$ and comparing them with reference solutions.
We will also test the abilities of DeepGenMSMs to generate physically
valid molecular configurations.

The networks were implemented using PyTorch \cite{paszke2017automatic}
and Tensorflow \cite{tensorflow2015-whitepaper_2}. For the full code
and all details about the neural network architecture, hyper-parameters
and training algorithm, please refer to \href{https://github.com/markovmodel/deep_gen_msm}{https://github.com/markovmodel/deep\_gen\_msm}. 

\subsection{Diffusion in Prinz potential}

\begin{figure}
\begin{centering}
\includegraphics[width=1\columnwidth]{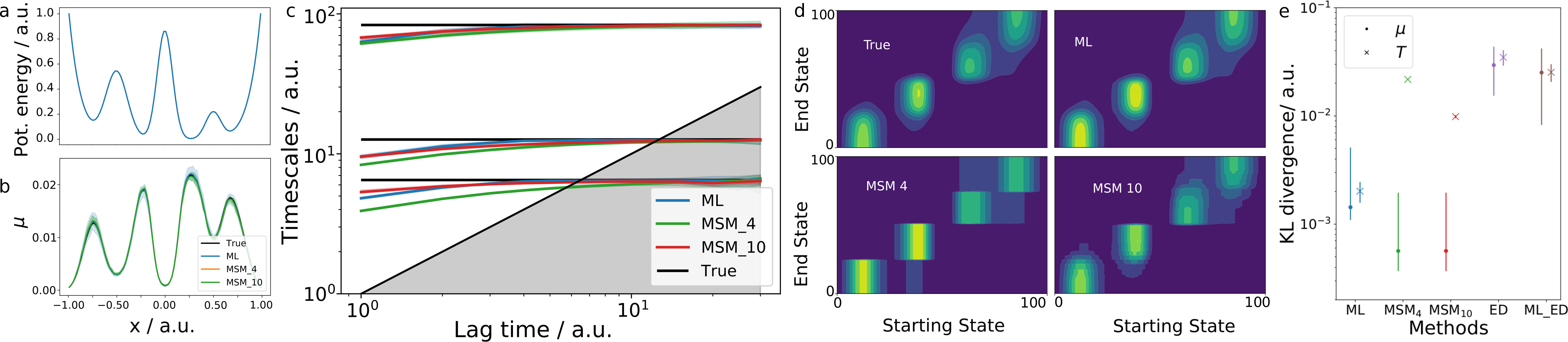}
\par\end{centering}
\caption{\label{fig:PrinzKinetics}Performance of deep versus standard MSMs
for diffusion in the Prinz Potential. (a) Potential energy as a function
of position $x$. (b) Stationary distribution estimates of all methods
with the exact distribution (black). (c) Implied timescales of the
Prinz potential compared to the real ones (black line). (d) True transition
density and approximations using maximum likelihood (ML) DeepResampleMSM,
four and ten state MSMs. (e) KL-divergence of the stationary and transition
distributions with respect to the true ones for all presented methods
(also DeepGenMSM).}
\end{figure}

We first apply our framework to the time-discretized diffusion process
$x_{t+\Delta t}=-\Delta t\,\nabla V(x_{t})+\sqrt{2\Delta t}\dot{\eta}_{t}$
with $\Delta t=0.01$ in the Prinz potential $V(x_{t})$ introduced
in \cite{PrinzEtAl_JCP10_MSM1} (Fig. \ref{fig:PrinzKinetics}a).
For this system we know exact results for benchmarking: the stationary
distribution and relaxation timescales (black lines in Fig. \ref{fig:PrinzKinetics}b,c)
and the transition density (Fig. \ref{fig:PrinzKinetics}d). We simulate
trajectories of lengths $250,000$ and $125,000$ time steps for training
and validation, respectively. For all methods, we repeat the data
generation and model estimation process 10 times and compute mean
and standard deviations for all quantities of interest, which thus
represent the mean and variance of the estimators.

The functions $\chi$, $\gamma$ and $G$ are represented with densely
connected neural networks. The details of the architecture and the
training procedure can be found in the Supplementary Information. 

We compare DeepResampleMSMs and DeepGenMSMs with standard MSMs using
four or ten states obtained with $k$-means clustering. Note that
standard MSMs do not directly operate on configuration space. When
using an MSM, the transition density (Eq. \ref{eq:Ansatz}) is thus
simulated by:
\[
x_{t}\overset{\chi(x_{t})}{\longrightarrow}i\overset{\sim\mathbf{K}_{i,*}}{\longrightarrow}j\overset{\sim\rho_{j}(y)}{\longrightarrow}x_{t+\tau},
\]
i.e., we find the cluster $i$ associated with a configuration $x_{t}$,
which is deterministic for regular MSMs, then sample the cluster $j$
at the next time-step, and sample from the conditional distribution
of configurations in cluster $j$ to generate $x_{t+\tau}$.

Both DeepResampleMSMs trained with the ML method and standard MSMs
can reproduce the stationary distribution within statistical uncertainty
(Fig. \ref{fig:PrinzKinetics}b). For long lag times $\tau$, all
methods converge from below to the correct relaxation timescales (Fig.
\ref{fig:PrinzKinetics}c), as expected from theory \cite{PrinzEtAl_JCP10_MSM1,NoeNueske_MMS13_VariationalApproach}.
When using equally many states (here: four), the DeepResampleMSM has
a much lower bias in the relaxation timescales than the standard MSM.
This is expected from approximation theory, as the DeepResampleMSMs
represents the four metastable states with a meaningful, smooth membership
functions $\boldsymbol{\chi}(x_{t})$, while the four-state MSM cuts
the memberships hard at boundaries with low sample density (Supplementary
Fig. 1). When increasing the number of metastable states, the bias
of all estimators will reduce. An MSM with ten states is needed to
perform approximately equal to a four-state DeepResampleMSM (Fig.
\ref{fig:PrinzKinetics}c). All subsequent analyses use a lag time
of $\tau=5$. 

The DeepResampleMSM generates a transition density that is very similar
to the exact density, while the MSM transition densities are coarse-grained
by virtue of the fact that $\boldsymbol{\chi}(x_{t})$ performs a
hard clustering in an MSM (Fig. \ref{fig:PrinzKinetics}d). This impression
is confirmed when computing the Kullback-Leibler divergence of the
distributions (Fig. \ref{fig:PrinzKinetics}e). 

\begin{figure}
\begin{centering}
\includegraphics[width=0.85\textwidth]{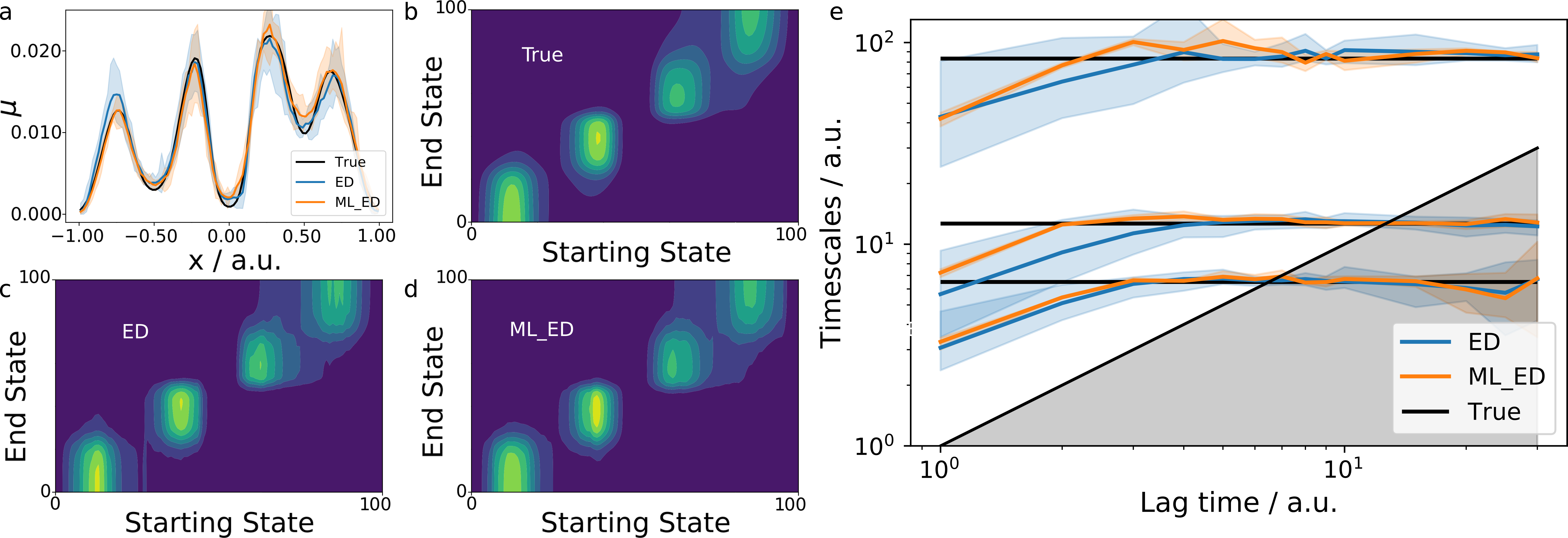}
\par\end{centering}
\caption{\label{fig:PrinzGenerative}Performance of DeepGenMSMs for diffusion
in the Prinz Potential. Comparison between exact reference (black),
DeepGenMSMs estimated using only energy distance (ED) or combined
ML-ED training. (a) Stationary distribution. (b-d) Transition densities.
(e) Relaxation timescales.}
\end{figure}

Encouraged by the accurate results of DeepResampleMSMs, we now train
DeepGenMSM, either by training both the $\boldsymbol{\chi}$ and $\mathbf{q}$
networks by minimizing the energy distance (ED), or by taking $\boldsymbol{\chi}$
from a ML-trained DeepResampleMSM and only training the $\mathbf{q}$
network by minimizing the energy distance (ML-ED). The stationary
densities, relaxation timescales and transition densities can still
be approximated in these settings, although the DeepGenMSMs exhibit
larger statistical fluctuations than the resampling MSMs (Fig. \ref{fig:PrinzGenerative}).
ML-ED appears to perform slightly better than ED alone, likely because
reusing $\boldsymbol{\chi}$ from the ML training makes the problem
of training the generator easier.

For a one-dimensional example like the Prinz potential, learning a
generative model does not provide any added value, as the distributions
can be well approximated by the empirical distributions. The fact
that we can still get approximately correct results for stationary,
kinetics and dynamical properties encourages us to use DeepGenMSMs
for a higher-dimensional example, where the generation of configurations
is a hard problem.

\subsection{Alanine dipeptide}

We use explicit-solvent MD simulations of Alanine dipeptide as a second
example. Our aim is the learn stationary and kinetic properties, but
especially to learn a generative model that generates genuinely novel
but physically meaningful configurations. One $\unit{\unit[250]{ns}}$
trajectory with a storage interval of $\unit[1]{ps}$ is used and
split $80\%/20\%$ for training and validation \textendash{} see \cite{MardtEtAl_VAMPnets}
for details of the simulation setup. We characterize all structures
by the three-dimensional Cartesian coordinates of the heavy atoms,
resulting in a 30 dimensional configuration space. While we do not
have exact results for Alanine dipeptide, the system is small enough
and well enough sampled, such that high-quality estimates of stationary
and kinetic properties can be obtained from a very fine MSM \cite{PrinzEtAl_JCP10_MSM1}.
We therefore define an MSM build on 400 equally sized grid areas in
the $(\phi,\psi)$-plane as a reference at a lag time of $\tau=\unit[25]{ps}$
that has been validated by established methods \cite{PrinzEtAl_JCP10_MSM1}.

Neural network and training details are again found at the git repository
and in the Supplementary Information. 

For comparison with deep MSMs, we build two standard MSMs following
a state of the art protocol: we transform input configurations with
a kinetic map preserving 95\% of the cumulative kinetic variance \cite{NoeClementi_JCTC15_KineticMap},
followed by $k$-means clustering, where $k=6$ and $k=100$ are used. 

\begin{figure}
\centering{}\includegraphics[width=1\columnwidth]{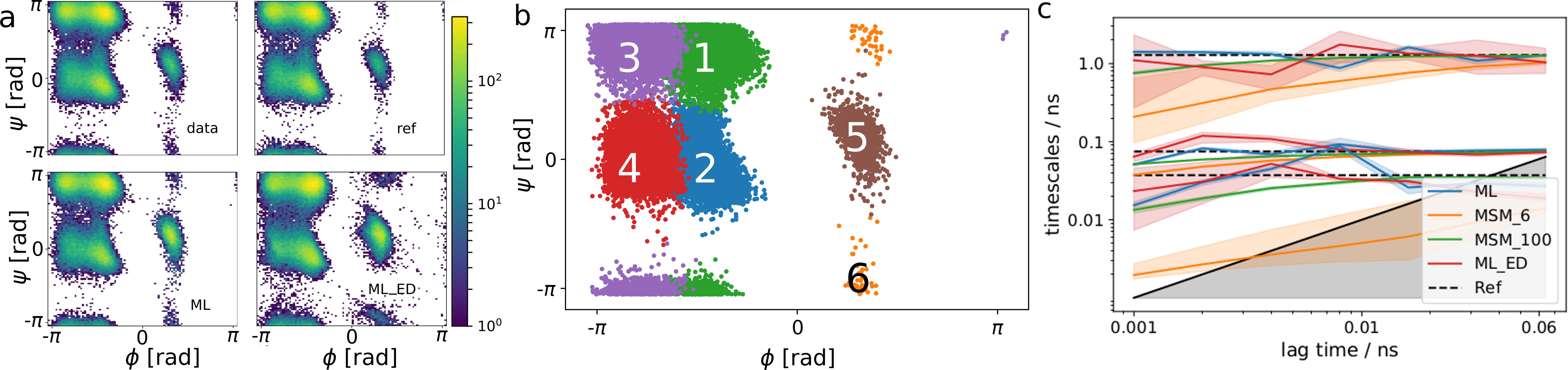}\caption{\label{fig:AlaKinetics}Performance of DeepResampleMSM and DeepGenMSMs
versus standard MSMs on the Alanine dipeptide simulation trajectory.
(a) Data distribution and stationary distributions from reference
MSM, DeepResampleMSM, and DeepGenMSM. (b) State classification by
DeepResampleMSM (c) Relaxation timescales.}
\end{figure}

DeepResampleMSM trained with ML method approximate the stationary
distribution very well (Fig. \ref{fig:AlaKinetics}a). The reference
MSM assigns a slightly lower weight to the lowest-populated state
6, but otherwise the data, reference distribution and deep MSM distribution
are visually indistinguishable. The relaxation timescales estimated
by a six-state DeepResampleMSM are significantly better than with
six-state standard MSMs. MSMs with 100 states have a similar performance
as the deep MSMs but this comes at the cost of a model with a much
larger latent space.

Finally, we test DeepGenMSMs for Alanine dipeptide where $\boldsymbol{\chi}$
is trained with the ML method and the generator is then trained using
ED (ML-ED). The stationary distribution generated by simulating the
DeepGenMSM recursively results in a stationary distribution which
is very similar to the reference distribution in states 1-4 with small
$\phi$ values (Fig. \ref{fig:AlaKinetics}a). States number 5 and
6 with large $\phi$ values are captured, but their shapes and weights
are somewhat distorted (Fig. \ref{fig:AlaKinetics}a). The one-step
transition densities predicted by the generator are high quality for
all states (Suppl. Fig. 2), thus the differences observed for the
stationary distribution must come from small errors made in the transitions
between metastable states that are very rarely observed for states
5 and 6. These rare events result in poor training data for the generator.
However, the DeepGenMSMs approximates the kinetics well within the
uncertainty that is mostly due to estimator variance (Fig. \ref{fig:AlaKinetics}c).

\begin{figure}
\begin{centering}
\includegraphics[width=0.9\columnwidth]{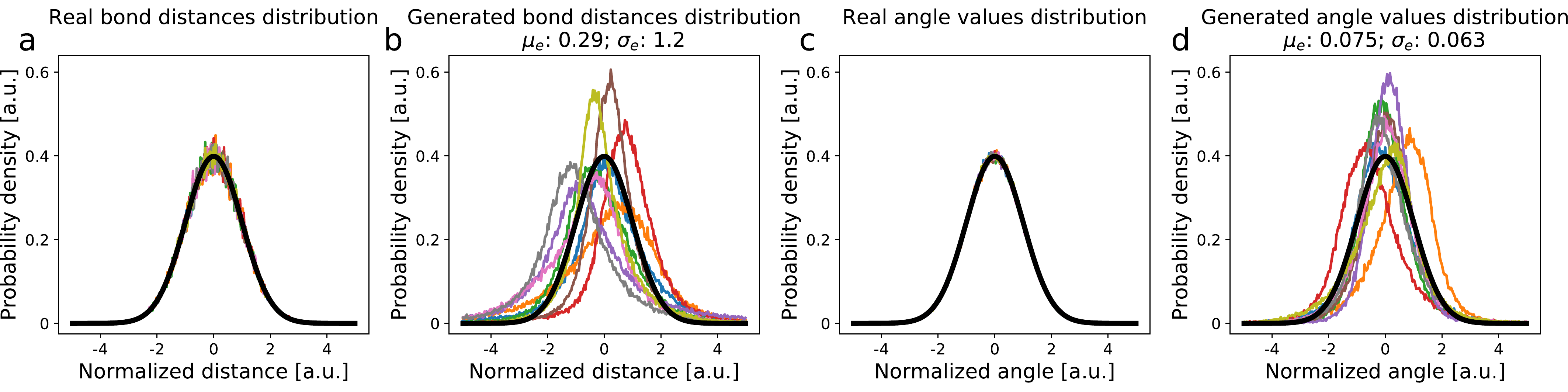}
\par\end{centering}
\caption{\label{fig:AlaBondDistributions}Normalized bond (a,b) and angle (c,d)
distributions of Alanine dipeptide compared to Gaussian normal distribution
(black). (a,c) True MD data. (b,d) Data from trajectories generated
by DeepGenMSMs.}
\end{figure}

Now we ask whether DeepGenMSMs can sample valid structures in the
30-dimensional configuration space, i.e., if the placement of atoms
is physically meaningful. As we generate configurations in Cartesian
space, we first check if the internal coordinates are physically viable
by comparing all bond lengths and angles between real MD data and
generated trajectories (Fig. \ref{fig:AlaBondDistributions}). The
true bond lengths and angles are almost perfectly Gaussian distributed,
and we thus normalize them by shifting each distribution to a mean
of 0 and scaling it to have standard deviation 1, which results all
reference distributions to collapse to a normal distribution (Fig.
\ref{fig:AlaBondDistributions}a,c). We normalize the generated distribution
with the mean and standard distribution of the true data. Although
there are clear differences (Fig. \ref{fig:AlaBondDistributions}b,d),
these distributions are very encouraging. Bonds and angles are very
stiff degrees of freedom, and the fact that most differences in mean
and standard deviation are small when compared to the true fluctuation
width means that the generated structures are close to physically
accurate and could be refined by little additional MD simulation effort.

Finally, we perform an experiment to test whether the DeepGenMSM is
able to generate genuinely new configurations that do exist for Alanine
dipeptide but have not been seen in the training data. In other words,
can the generator ``extrapolate'' in a meaningful way? This is a
fundamental question, because simulating MD is exorbitantly expensive,
with each simulation time step being computationally expensive but
progressing time only of the order of $10^{-15}$ seconds, while often
total simulation timescales of $10^{-3}$ seconds or longer are needed.
A DeepGenMSM that makes leaps of length $\tau$ \textendash{} orders
of magnitude larger than the MD simulation time-step \textendash{}
and has even a small chance of generating new and meaningful structures
would be extremely valuable to discover new states and thereby accelerate
MD sampling.

\begin{figure}
\begin{centering}
\includegraphics[width=1\columnwidth]{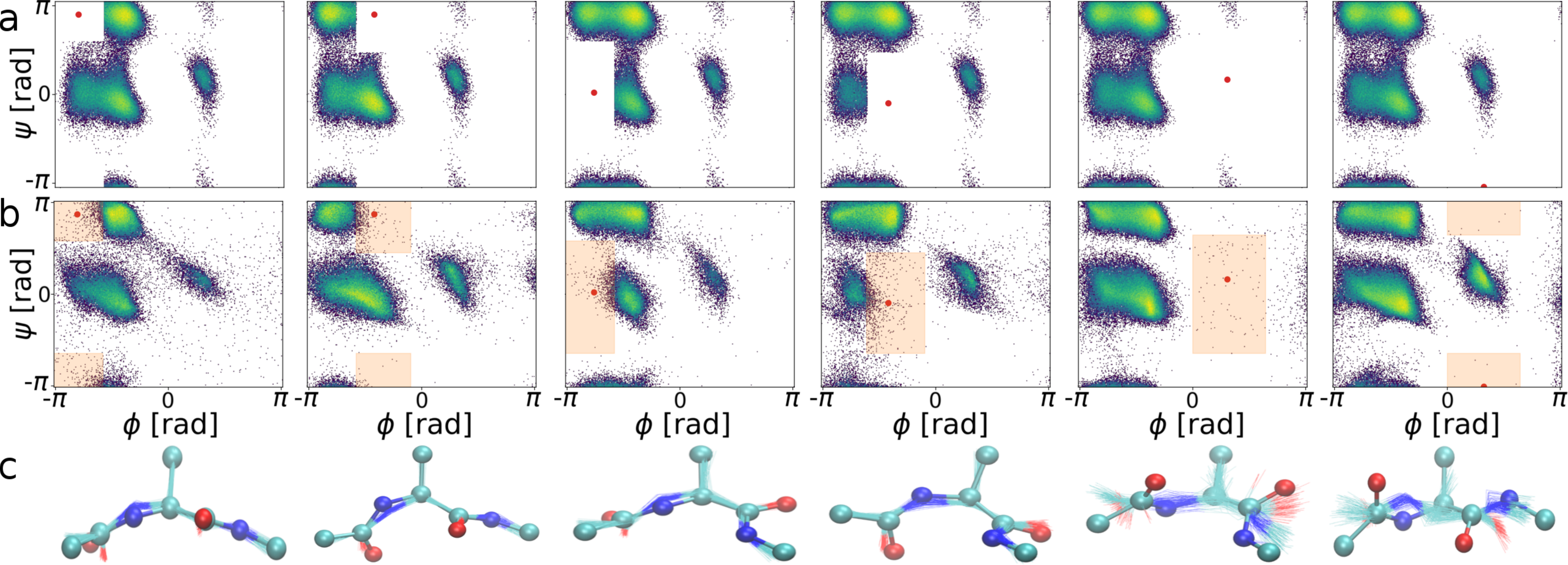}
\par\end{centering}
\caption{\label{fig:AlaGenerative}DeepGenMSMs can generate physically realistic
structures in areas that were not included in the training data. (a)
Distribution of training data. (b) Generated stationary distribution.
(c) Representative ``real'' molecular configuration (from MD simulation)
in each of the metastable states (sticks and balls), and the 100 closest
configurations generated by the DeepGenMSM (lines).}
\end{figure}

To test this ability, we conduct six experiments, in each of which
we remove all data belonging to one of the six metastable states of
Alanine dipeptide (\ref{fig:AlaGenerative}a). We train a DeepGenMSM
with each of these datasets separately, and simulate it to predict
the stationary distribution (\ref{fig:AlaGenerative}b). While the
generated stationary distributions are skewed and the shape of the
distribution in the $(\phi,\psi)$ range with missing-data are not
quantitatively predicted, the DeepGenMSMs do indeed predict configurations
where no training data was present (\ref{fig:AlaGenerative}b). Surprisingly,
the quality of most of these configurations is high (\ref{fig:AlaGenerative}c).
While the structures of the two low-populated states 5-6 do not look
realistic, each of the metastable states 1-4 are generated with high
quality, as shown by the overlap of a real MD structure and the 100
most similar generated structures (\ref{fig:AlaGenerative}c).

In conclusion, deep MSMs provide high-quality models of the stationary
and kinetic properties for stochastic dynamical systems such as MD
simulations. In contrast to other high-quality models such as VAMPnets,
the resulting model is truly probabilistic and can thus be physically
interpreted and be used in a Bayesian framework. For the first time,
it was shown that generating dynamical trajectories in a 30-dimensional
molecular configuration space results in sampling of physically realistic
molecular structures. While Alanine dipeptide is a small system compared
to proteins and other macromolecules that are of biological interest,
our results demonstrate that efficient sampling of new molecular structures
is possible with generative dynamic models, and improved methods can
be built upon this. Future methods will especially need to address
the difficulties of generating valid configurations in low-probability
regimes, and it is likely that the energy distance used here for generator
training needs to be revisited to achieve this goal.

\paragraph{Acknowledgements}

This work was funded by the European Research Commission (ERC CoG
``ScaleCell''), Deutsche Forschungsgemeinschaft (CRC 1114/A04, Transregio
186/A12, NO 825/4--1, Dynlon P8), and the ``1000-Talent Program
of Young Scientists in China''.

\bibliographystyle{plain}
\bibliography{all,references,own}

\begin{thebibliography}{10}

\bibitem{arjovsky2017wasserstein}
Arjovsky, Martin and Chintala, Soumith and Bottou, L{\'e}on.
\newblock {Wasserstein generative adversarial networks}.
\newblock {\em International Conference on Machine Learning}, 214--223, 2017.

\bibitem{BowmanEnsignPande_JCTC2010_AdaptiveSampling}
G.~R. Bowman, D.~L. Ensign, and V.~S. Pande.
\newblock {Enhanced Modeling via Network Theory: Adaptive Sampling of Markov
  State Models}.
\newblock {\em J. Chem. Theory Comput.}, 6(3):787--794, 2010.

\bibitem{BowmanPandeNoe_MSMBook}
G.~R. Bowman, V.~S. Pande, and F.~No{\'e}.
\newblock {\em An Introduction to Markov State Models and Their Application to
  Long Timescale Molecular Simulation}.
\newblock 2014.

\bibitem{BruntonProctorKutz_PNAS16_Sindy}
S.~L. Brunton, J.~L. Proctor, and J.~N. Kutz.
\newblock Discovering governing equations from data by sparse identification of
  nonlinear dynamical systems.
\newblock {\em Proc. Natl. Acad. Sci. USAP}, 113:3932--3937.

\bibitem{clevert2015fast}
Djork-Arn{\'e} Clevert, Thomas Unterthiner, and Sepp Hochreiter.
\newblock Fast and accurate deep network learning by exponential linear units
  (elus).
\newblock {\em arXiv preprint arXiv:1511.07289}, 2015.

\bibitem{tensorflow2015-whitepaper_2}
Mart\'{\i}n~Abadi et~al.
\newblock {TensorFlow}: Large-scale machine learning on heterogeneous systems,
  2015.
\newblock Software available from tensorflow.org.

\bibitem{he2016deep}
Kaiming He, Xiangyu Zhang, Shaoqing Ren, and Jian Sun.
\newblock Deep residual learning for image recognition.
\newblock In {\em Proceedings of the IEEE conference on computer vision and
  pattern recognition}, pages 770--778, 2016.

\bibitem{HernandezPande_VariationalDynamicsEncoder}
C.~X. Hern{\'a}ndez, H.~K. Wayment-Steele, M.~M. Sultan, B.~E. Husic, and V.~S.
  Pande.
\newblock Variational encoding of complex dynamics.
\newblock {\em arXiv:1711.08576}, 2017.

\bibitem{ioffe2015batch}
Sergey Ioffe and Christian Szegedy.
\newblock Batch normalization: Accelerating deep network training by reducing
  internal covariate shift.
\newblock {\em arXiv preprint arXiv:1502.03167}, 2015.

\bibitem{adam}
Diederik~P. Kingma and Jimmy Ba.
\newblock Adam: {A} method for stochastic optimization.
\newblock {\em CoRR}, abs/1412.6980, 2014.

\bibitem{KordaMezic_JNLS2017_ConvergenceEDMD}
Milan Korda and Igor Mezic.
\newblock On convergence of extended dynamic mode decomposition to the koopman
  operator.
\newblock {\em J. Nonlinear Sci.}, 28:687--710, 2017.

\bibitem{KubeWeber_JCP07_CoarseGraining}
S.~Kube and M.~Weber.
\newblock {A coarse graining method for the identification of transition rates
  between molecular conformations}.
\newblock {\em J. Chem. Phys.}, 126:024103, 2007.

\bibitem{LiKevekidis_EDMD-DL}
Q.~Li, F.~Dietrich, E.~M. Bollt, and I.~G. Kevrekidis.
\newblock Extended dynamic mode decomposition with dictionary learning: a
  data-driven adaptive spectral decomposition of the koopman operator.
\newblock {\em Chaos}, 27:103111, 2017.

\bibitem{LuschKutzBrunton_DeepKoopman}
B.~Lusch and S.~L.~Brunton J~. N.~Kutz.
\newblock Deep learning for universal linear embeddings of nonlinear dynamics.
\newblock {\em arXiv:1712.09707}, 2017.

\bibitem{MardtEtAl_VAMPnets}
Andreas Mardt, Luca Pasquali, Hao Wu, and Frank No{\'e}.
\newblock Vampnets for deep learning of molecular kinetics.
\newblock {\em Nat. Commun.}, 9(1):5, 2018.

\bibitem{Mezic_NonlinDyn05_Koopman}
I.~Mezi\'{c}.
\newblock Spectral properties of dynamical systems, model reduction and
  decompositions.
\newblock {\em Nonlinear Dynam.}, 41:309--325, 2005.

\bibitem{NoeClementi_JCTC15_KineticMap}
Frank No{\'e} and Cecilia Clementi.
\newblock Kinetic distance and kinetic maps from molecular dynamics simulation.
\newblock {\em J. Chem. Theory Comput.}, 11(10):5002--5011, 2015.

\bibitem{NoeNueske_MMS13_VariationalApproach}
Frank No{\'e} and Feliks Nuske.
\newblock A variational approach to modeling slow processes in stochastic
  dynamical systems.
\newblock {\em Multiscale Model. \& Simul.}, 11(2):635--655, 2013.

\bibitem{OttoRowley_LinearlyRecurrentAutoencoder}
S.~E. Otto and C.~W. Rowley.
\newblock Linearly-recurrent autoencoder networks for learning dynamics.
\newblock {\em arXiv:1712.01378}, 2017.

\bibitem{paszke2017automatic}
Adam Paszke, Sam Gross, Soumith Chintala, Gregory Chanan, Edward Yang, Zachary
  DeVito, Zeming Lin, Alban Desmaison, Luca Antiga, and Adam Lerer.
\newblock Automatic differentiation in pytorch.
\newblock In {\em NIPS-W}, 2017.

\bibitem{PlattnerEtAl_NatChem17_BarBar}
Nuria Plattner, Stefan Doerr, Gianni De~Fabritiis, and Frank No{\'e}.
\newblock Complete protein--protein association kinetics in atomic detail
  revealed by molecular dynamics simulations and markov modelling.
\newblock {\em Nat. Chem.}, 9(10):1005, 2017.

\bibitem{PrinzEtAl_JCP10_MSM1}
Jan-Hendrik Prinz, Hao Wu, Marco Sarich, Bettina Keller, Martin Senne, Martin
  Held, John~D Chodera, Christof Sch{\"u}tte, and Frank No{\'e}.
\newblock Markov models of molecular kinetics: Generation and validation.
\newblock {\em J. Chem. Phys.}, 134(17):174105, 2011.

\bibitem{RibeiroTiwary_JCP18_RAVE}
Jo{\~a}o Marcelo~Lamim Ribeiro, Pablo Bravo, Yihang Wang, and Pratyush Tiwary.
\newblock Reweighted autoencoded variational bayes for enhanced sampling
  (rave).
\newblock {\em J. Chem. Phys.}, 149:072301, 2018.

\bibitem{SarichSchuette_MSMBook13}
M.~Sarich and C.~Sch\"{u}tte.
\newblock {\em Metastability and Markov State Models in Molecular Dynamics}.
\newblock Courant Lecture Notes. American Mathematical Society, 2013.

\bibitem{SarichNoeSchuette_MMS09_MSMerror}
Marco Sarich, Frank No{\'e}, and Christof Sch{\"u}tte.
\newblock On the approximation quality of markov state models.
\newblock {\em Multiscale Model. Simul.}, 8(4):1154--1177, 2010.

\bibitem{SchmidSesterhenn_APS08_DMD}
P.~J. Schmid and J.~Sesterhenn.
\newblock Dynamic mode decomposition of numerical and experimental data.
\newblock In {\em 61st Annual Meeting of the APS Division of Fluid Dynamics.
  American Physical Society}, 2008.

\bibitem{SwopePiteraSuits_JPCB108_6571}
W.~C. Swope, J.~W. Pitera, and F.~Suits.
\newblock {Describing protein folding kinetics by molecular dynamics
  simulations: 1. Theory}.
\newblock {\em J. Phys. Chem. B}, 108:6571--6581, 2004.

\bibitem{SzekelyRizzo_Interstat04_EnergyDistance}
G.~Sz{\'e}kely and M.~Rizzo.
\newblock Testing for equal distributions in high dimension.
\newblock {\em InterStat,}, 5, 2004.

\bibitem{TuEtAl_JCD14_ExactDMD}
J.~H. Tu, C.~W. Rowley, D.~M. Luchtenburg, S.~L. Brunton, and J.~N. Kutz.
\newblock On dynamic mode decomposition: Theory and applications.
\newblock {\em J. Comput. Dyn.}, 1(2):391--421, dec 2014.

\bibitem{WehmeyerNoe_TAE}
Christoph Wehmeyer and Frank No{\'e}.
\newblock Time-lagged autoencoders: Deep learning of slow collective variables
  for molecular kinetics.
\newblock {\em J. Chem. Phys.}, 148(24):241703, 2018.

\bibitem{WuNoe_VAMP}
Hao Wu and Frank No{\'e}.
\newblock Variational approach for learning markov processes from time series
  data.
\newblock {\em arXiv:1707.04659}, 2017.

\end{thebibliography}

\clearpage

\section*{Supplementary Material}

\setcounter{figure}{0}
\renewcommand{\figurename}{Supplementary Fig.}

\subsection*{Analysis of transition matrices}

The transition matrix $\mathbf{K}(\tau)=[k_{ij}(\tau)]\in\mathbb{R}^{m\times m}$
is defined as
\[
k_{ij}(\tau)=\mathbb{P}(x_{t+\tau}\in\text{state }j|x_{t}\in\text{state }i).
\]
Then, according to (\ref{eq:Ansatz}), we have
\begin{eqnarray*}
k_{ij}(\tau) & = & \int\mathbb{P}(x_{t+\tau}=y|x_{t}\in\text{state }i)\cdot\mathbb{P}(x_{t+\tau}\in\text{state }j|x_{t+\tau}=y)\mathrm{d}y\\
 & = & \int q_{i}(y;\tau)\chi_{j}(y)\,\mathrm{d}y,
\end{eqnarray*}
and
\[
k_{ij}(n\tau)=\left[\mathbf{K}(\tau)^{n}\right]_{ij}.
\]

If $\boldsymbol{\chi},\mathbf{q}$ satisfy the conditions
\[
\begin{array}{ll}
\chi_{i}(x)\ge0,\sum_{j}\chi_{j}(x)=1,\\
q_{i}(x;\tau)\ge0,\int q_{i}(y;\tau)\mathrm{d}y=1, & \forall x,i,
\end{array}
\]
we have $k_{ij}(\tau)\ge0$ and $\sum_{j}k_{ij}(\tau)=1$, i.e., $\mathbf{K}(\tau)$
computed from $\boldsymbol{\chi},\mathbf{q}$ is a valid transition
probability matrix.

For the distribution $\mu$ defined in (\ref{eq:stationary_density}),
\begin{eqnarray*}
\int\mathbb{P}(x_{t+\tau}=y|x_{t}=x)\cdot\mu(x)\mathrm{d}x & = & \int\mathbf{q}(y;\tau)^{\top}\boldsymbol{\chi}(x)\cdot\mathbf{q}(x;\tau)^{\top}\boldsymbol{\pi}\mathrm{d}x\\
 & = & \mathbf{q}(y;\tau)^{\top}\boldsymbol{\pi}\\
 & = & \mu(y),
\end{eqnarray*}
which shows $\mu$ is the stationary distribution of model (\ref{eq:Ansatz}).

\subsection*{VAMP-E training of deep MSMs}

An alternative to ML training is to employ a score from the Variational
Approach of Markov Processes (VAMP) \cite{WuNoe_VAMP}. The VAMP-E
score has the advantage over other VAMP scores employed previously
\cite{MardtEtAl_VAMPnets} that we do not have to specify the rank
of the model \cite{WuNoe_VAMP}. The VAMP-E score is computed as

\begin{equation}
\mathcal{R}_{E}=\mathrm{tr}\left(2\mathbf{C}_{01}\bar{\boldsymbol{\Gamma}}^{-1}-\mathbf{C}_{00}\bar{\boldsymbol{\Gamma}}^{-1}\mathbf{C}_{11}\bar{\boldsymbol{\Gamma}}^{-1}\right)\label{eq:VE-1}
\end{equation}
which depends on covariance matrices estimated from the transformed
data: 
\begin{eqnarray*}
[\mathbf{C}_{00}]_{ij} & = & \mathbb{E}_{t}[\chi_{i}(x_{t})\chi_{j}(x_{t})]\\{}
[\mathbf{C}_{11}]_{ij} & = & \mathbb{E}_{t}[\gamma_{i}(x_{t+\tau})\gamma_{j}(x_{t+\tau})]\\{}
[\mathbf{C}_{01}]_{ij} & = & \mathbb{E}_{t}[\chi_{i}(x_{t})\gamma_{j}(x_{t+\tau})]\\
\bar{\boldsymbol{\Gamma}} & = & \mathrm{diag}(\bar{\gamma}_{1},\ldots,\bar{\gamma}_{m}).
\end{eqnarray*}
We can use the standard empirical estimators to compute $\mathbb{E}_{t}$.
We can then train a deep MSM using the structure shown in Fig. \ref{fig:scheme}
by maximizing (\ref{eq:VE-1}).

\subsection*{Using the Energy Distance to train generative networks}

The Energy Distance (ED) \cite{SzekelyRizzo_Interstat04_EnergyDistance}
is a metric that measures the difference between the distributions
of two real valued random vectors $x$ and $y$, and is defined as
\begin{equation}
D_{E}(\mathbb{P}(x),\mathbb{P}(y))=\mathbb{E}\left[2\left\Vert x-y\right\Vert -\left\Vert x-x^{\prime}\right\Vert -\left\Vert y-y^{\prime}\right\Vert \right].
\end{equation}
Here, $x^{\prime},y^{\prime}$ are independently distributed according
to the distributions of $y,z$. Therefore, the conditional energy
distance given in (\ref{eq:conditional-energy-distance}) is equal
to the mean value of the energy distance between the conditional distributions
$\mathbb{P}(x_{t+\tau}|x_{t})$ and $\mathbb{P}(\hat{x}_{t+\tau}|x_{t})$
for all $x_{t}$, and satisfies that $D\ge0$ and $D=0$ if and only
if $\mathbb{P}(x_{t+\tau}|x_{t})=\mathbb{P}(\hat{x}_{t+\tau}|x_{t})$
for all $x_{t}$.

Noticing that $\mathbb{E}\left[\left\Vert x_{t+\tau}-x_{t+\tau}^{\prime}\right\Vert \right]$
is a constant for a given system. We can therefore approximate $D$
as
\begin{eqnarray*}
D & = & \mathbb{E}\left[\left\Vert \hat{x}_{t+\tau}-x_{t+\tau}\right\Vert +\left\Vert \hat{x}_{t+\tau}^{\prime}-x_{t+\tau}\right\Vert -\left\Vert \hat{x}_{t+\tau}-\hat{x}_{t+\tau}^{\prime}\right\Vert \right]+\mathrm{const}\\
 & = & \mathbb{E}[d_{t}]+\mathrm{const}\\
 & \approx & \frac{1}{N}\sum_{t}d_{t}+\mathrm{const}
\end{eqnarray*}
where
\begin{equation}
d_{t}=\left\Vert G(e_{I_{t}},\epsilon_{t})-x_{t+\tau}\right\Vert +\left\Vert G(e_{I_{t}^{\prime}},\epsilon_{t}^{\prime})-x_{t+\tau}\right\Vert -\left\Vert G(e_{I_{t}},\epsilon_{t})-G(e_{I_{t}^{\prime}},\epsilon_{t}^{\prime})\right\Vert 
\end{equation}
Here, $N=T-\tau$ is the number of all transition pairs $(x_{t},x_{t+\tau})$
present in the trajectory data, $I_{t},I_{t}^{\prime}$ are discrete
random variables with $\mathbb{P}(I_{t}=i)=\mathbb{P}(I_{t}^{\prime}=i)=\chi_{i}(x_{t})$,
and $\epsilon_{t},\epsilon_{t}^{\prime}$ are i.i.d random vectors
whose components have Gaussian normal distributions.

The gradient of $D$ with respect to parameters $W_{G}$ of the generative
model $G$ can be unbiasedly estimated by the mean value of $\partial d_{t}/\partial W_{G}$.
But for parameters $W_{\chi}$ of $\boldsymbol{\chi}$, $\partial d_{t}/\partial W_{\chi}$
does not exist because $I_{t},I_{t}^{\prime}$ is discrete-valued.
In order to overcome this problem, we assume here $\boldsymbol{\chi}(x)=\mathrm{SoftMax}\left[\mathbf{o}(x)\right]$
is modeled by a neural network with the softmax output layer. Then
\begin{eqnarray*}
\frac{\partial}{\partial o_{k}}\mathbb{E}[d_{t}|x_{t},x_{t+\tau}] & = & \sum_{i,j}\chi_{i}(x)\chi_{j}(x)\left(1_{i=k}+1_{j=k}-2\chi_{k}(x_{t})\right)\\
 &  & \cdot\mathbb{E}\left[\left\Vert G(e_{i},\epsilon_{t})-x_{t+\tau}\right\Vert +\left\Vert G(e_{j},\epsilon_{t}^{\prime})-x_{t+\tau}\right\Vert -\left\Vert G(e_{i},\epsilon_{t})-G(e_{j},\epsilon_{t}^{\prime})\right\Vert \right]\\
 & = & \mathbb{E}\left[\left(1_{I_{t}=k}+1_{I_{t}^{\prime}=k}-2\chi_{k}(x_{t})\right)\cdot d_{t}\right],
\end{eqnarray*}
which leads to the estimation
\begin{eqnarray*}
\frac{\partial D}{\partial W_{\chi}} & = & \sum_{k}\frac{\partial o_{k}}{\partial W_{\chi}}\frac{\partial D}{\partial o_{k}}\\
 & \approx & \frac{1}{N}\sum_{t}d_{t}\sum_{k}\left(1_{I_{t}=k}+1_{I_{t}^{\prime}=k}-2\chi_{k}(x_{t})\right)\frac{\partial o_{k}}{\partial W_{\chi}}.
\end{eqnarray*}
By using the stochastic gradient over mini-batch over the entire data,
we can train the generative MSM as follows the subsequent algorithm:
\begin{enumerate}
\item Randomly choose a mini-batch $\{(x_{(n)},y_{(n)})\}_{i=1}^{B}\subset\{(x_{t},x_{t+\tau})\}$
with batch size $B$.
\item Draw $I_{(n)},I_{(n)}^{\prime}$ with
\begin{equation}
\mathbb{P}(I_{(n)}=i)=\mathbb{P}(I_{(n)}^{\prime}=i)=\chi_{i}(\tilde{x}_{(n)}),
\end{equation}
and draw $\epsilon_{(n)},\epsilon_{(n)}^{\prime}$ according to the
Gaussian distribution for $n=1,\ldots,B$.
\item Compute
\begin{eqnarray}
\delta W_{G} & = & \frac{1}{B}\sum_{n=1}^{B}\frac{\partial d_{(n)}}{\partial W_{G}}\nonumber \\
\delta W_{\chi} & = & \frac{1}{B}\sum_{i=1}^{B}d_{(n)}\cdot\sum_{k=1}^{m}\left(1_{I_{(n)}=k}+1_{I_{(n)}^{\prime}=k}-2\chi_{k}(x_{(n)})\right)\frac{\partial o_{k}(x_{(n)})}{\partial W_{\chi}}
\end{eqnarray}
with
\begin{equation}
d_{(n)}=\left\Vert G(e_{I_{(n)}},\epsilon_{(n)})-y_{(n)}\right\Vert +\left\Vert G(e_{I_{(n)}^{\prime}},\epsilon_{(n)}^{\prime})-y_{(n)}\right\Vert -\left\Vert G(e_{I_{(n)}},\epsilon_{(n)})-G(e_{I_{(n)}^{\prime}},\epsilon_{(n)}^{\prime})\right\Vert 
\end{equation}
and $\boldsymbol{\chi}=\mathrm{SoftMax}\left[\mathbf{o}\right]$.
\item Update
\begin{align*}
W_{G} & \leftarrow W_{G}-\eta\delta W_{G}\\
W_{P} & \leftarrow W_{P}-\eta\delta W_{P}
\end{align*}
with a learning rate $\eta$.
\end{enumerate}

\subsection*{Motivation of Energy Distance as the training metric}

The major advantages of ED are:
\begin{enumerate}
\item It can be unbiasedly estimated from the data without an extra \textquotedblleft adversarial\textquotedblright{}
network as in GANs.
\item Unlike the KL divergence (see example 1 in \cite{arjovsky2017wasserstein}),
ED does not diverge in the case of few data points (small batch sizes)
or low populated probability density areas.
\item As a specific Maximum Mean Discrepancy (MMD), ED can avoid the problem
of popular kernel-MMDs that the gradients of cost functions are vanished
if the generated samples are far away from the training data, and
therefore achieve higher efficiency when learning generative models.
\end{enumerate}

\subsection*{Network architecture and training procedure\label{subsec:Network-architecture-and}}

All neural networks representing the functions $\chi$, $\gamma$
and $G$ for the Prinz potential are using 64 nodes in all 4 hidden
layers and batch normalization after each layer \cite{ioffe2015batch}.
Rectified linear activation functions (ReLUs) are used, except for
the output layer of $\chi$ which uses SoftMax and the output layer
of $G$ which has a linear activation function. Both $\chi$ and $\gamma$
have 4 output nodes, and $G$ receives a four-dimensional 1-hot-encoding
of the metastable state plus a four-dimensional noise vector as inputs.
Optimization is done using Adam \cite{adam}, with early stopping
checking if the validation score is not increasing over 5 epochs.
The learning rate for the training of $\chi$, $\gamma$ is $\lambda=10^{-3},$
and for $G$ $\lambda=10^{-5}$ with a batchsize of $100$. We are
using a time-lag of $\tau=5$ frames.

For alanine dipeptide, $\chi$ and $\gamma$ consist both of 3 residual
blocks \cite{he2016deep} built of 3 layers all having 100 nodes,
with exponential linear units (ELUs) \cite{clevert2015fast}, and
batch normalization for each layer. The output layer has 6 output
nodes, where $\chi$ uses a softmax activation function and $\gamma$
a RELU, respectively. In order to find all slow processes, it was
necessary to pre-train $\chi$ with the VAMPnet method \cite{MardtEtAl_VAMPnets}.
The generator $G$ uses 6 noise inputs and a six-dimensional 1-hot-encoding
of the metastable state and the ML-ED scheme. Networks are trained
with Adam until the validation score converges with a learning rate
of $\lambda=10^{-5}$ for $\chi$, $\gamma$ using $8000$ as batchsize
and $\lambda=10^{-4}$ for $G$ using $1500$ frames for a batch.
All subsequent analyses that use a fixed lag time employ $\tau=\unit[1]{ps}$. 

For finding the hyperparameter we performed a restricted grid search,
which showed comparing the KL divergence between the modeled distributions
that the result does only marginally depend on the choice of the parameters
(see \ref{tab:Hyperparameter-comparison-of} for an example).

\begin{table}
\centering%
\begin{tabular}{cccc}
\hline 
depth & width & dim random & KL div. / $10^{-2}$\tabularnewline
\hline 
\hline 
2 & 16 & 1 & $1.7$\tabularnewline
\hline 
2 & 16 & 2 & $2.2$\tabularnewline
\hline 
2 & 16 & 4 & $2.3$\tabularnewline
\hline 
2 & 32 & 1 & $2.2$\tabularnewline
\hline 
2 & 32 & 2 & $2.2$\tabularnewline
\hline 
2 & 32 & 4 & $2.5$\tabularnewline
\hline 
2 & 64 & 1 & $2.4$\tabularnewline
\hline 
2 & 64 & 2 & $2.7$\tabularnewline
\hline 
2 & 64 & 4 & $2.8$\tabularnewline
\hline 
2 & 128 & 1 & $2.7$\tabularnewline
\hline 
2 & 128 & 2 & $3.2$\tabularnewline
\hline 
2 & 128 & 4 & $3.8$\tabularnewline
\hline 
4 & 16 & 1 & $2.0$\tabularnewline
\hline 
4 & 16 & 2 & $1.6$\tabularnewline
\hline 
4 & 16 & 4 & $2.8$\tabularnewline
\hline 
4 & 32 & 1 & $1.7$\tabularnewline
\hline 
4 & 32 & 2 & $3.4$\tabularnewline
\hline 
4 & 32 & 4 & $3.2$\tabularnewline
\hline 
4 & 64 & 1 & $2.3$\tabularnewline
\hline 
4 & 64 & 2 & $2.9$\tabularnewline
\hline 
4 & 64 & 4 & $3.5$\tabularnewline
\hline 
4 & 128 & 1 & $1.8$\tabularnewline
\hline 
4 & 128 & 2 & $2.8$\tabularnewline
\hline 
4 & 128 & 4 & $1.8$\tabularnewline
\hline 
6 & 16 & 1 & $1.5$\tabularnewline
\hline 
6 & 16 & 2 & $3.5$\tabularnewline
\hline 
6 & 16 & 4 & $2.5$\tabularnewline
\hline 
6 & 32 & 1 & $3.3$\tabularnewline
\hline 
6 & 32 & 2 & $1.9$\tabularnewline
\hline 
6 & 32 & 4 & $2.3$\tabularnewline
\hline 
6 & 64 & 1 & $1.5$\tabularnewline
\hline 
6 & 64 & 2 & $2.7$\tabularnewline
\hline 
6 & 64 & 4 & $2.5$\tabularnewline
\hline 
6 & 128 & 1 & $1.6$\tabularnewline
\hline 
6 & 128 & 2 & $3.1$\tabularnewline
\hline 
6 & 128 & 4 & $1.7$\tabularnewline
\hline 
8 & 16 & 1 & $1.8$\tabularnewline
\hline 
8 & 16 & 2 & $1.9$\tabularnewline
\hline 
8 & 16 & 4 & $2.0$\tabularnewline
\hline 
8 & 32 & 1 & $1.6$\tabularnewline
\hline 
8 & 32 & 2 & $2.2$\tabularnewline
\hline 
8 & 32 & 4 & $2.7$\tabularnewline
\hline 
8 & 64 & 1 & $1.2$\tabularnewline
\hline 
8 & 64 & 2 & $2.4$\tabularnewline
\hline 
8 & 64 & 4 & $2.2$\tabularnewline
\hline 
8 & 128 & 1 & $1.8$\tabularnewline
\hline 
8 & 128 & 2 & $2.0$\tabularnewline
\hline 
8 & 128 & 4 & $1.6$\tabularnewline
\hline 
\end{tabular}

\caption{\label{tab:Hyperparameter-comparison-of}Hyperparameter comparison
of the KL divergence of the generated stationary distribution with
respect to the true one for the Prinz potential varying the depth,
the width, and the random input dimension taking the mean over $5$
runs.}

\end{table}

\subsection*{Supplementary Figures}

\begin{figure}[H]
\centering{}\includegraphics[width=1\columnwidth]{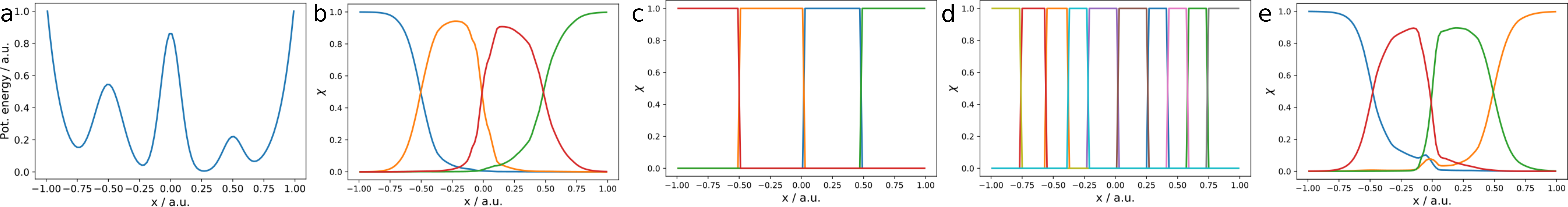}\caption{$\boldsymbol{\chi}(x)$ of the Prinz potential (a) Potential energy
as a function of position x. (b) Maximum Likelihood (c) four state
MSM (d) 10 state MSM (e) energy distance.}
\end{figure}

\begin{figure}[H]
\centering{}\includegraphics[width=1\columnwidth]{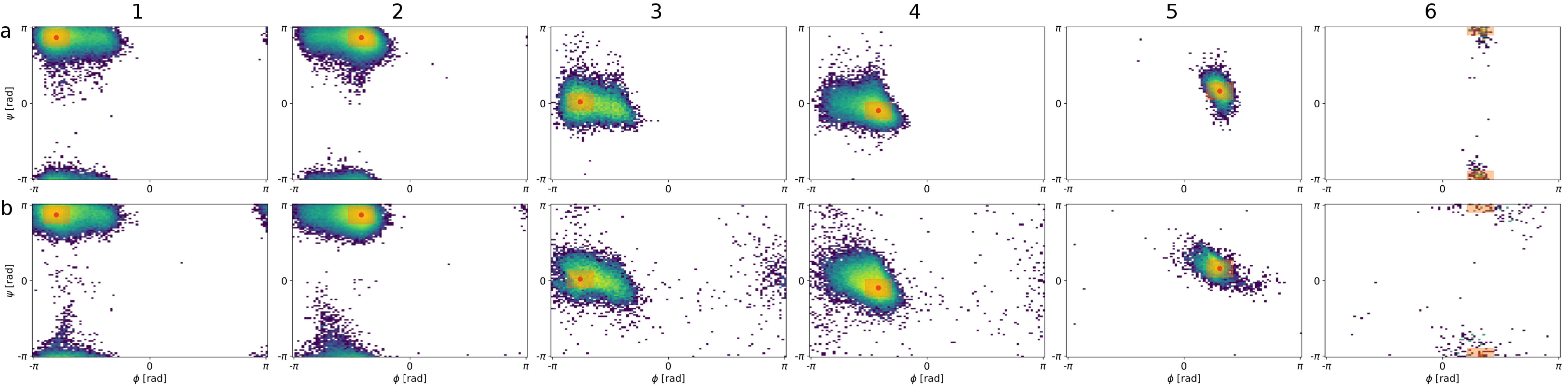}\caption{Conditional transition distributions for Alanine dipeptide starting
from different metastable states. The starting distribution are sampled
from the empirical distribution in the yellow region around the red
point. (a) Distribution sampled from the MD simulation. (b) Distribution
generated by the DeepGenMSM.}
\end{figure}

\end{document}